\def\year{2022}\relax
\documentclass[letterpaper]{article} % DO NOT CHANGE THIS
\usepackage{aaai22}  % DO NOT CHANGE THIS
\usepackage{times}  % DO NOT CHANGE THIS
\usepackage{helvet}  % DO NOT CHANGE THIS
\usepackage{courier}  % DO NOT CHANGE THIS
\usepackage[hyphens]{url}  % DO NOT CHANGE THIS
\usepackage{graphicx} % DO NOT CHANGE THIS
\usepackage{natbib}  % DO NOT CHANGE THIS AND DO NOT ADD ANY OPTIONS TO IT
\usepackage{caption} % DO NOT CHANGE THIS AND DO NOT ADD ANY OPTIONS TO IT
\DeclareCaptionStyle{ruled}{labelfont=normalfont,labelsep=colon,strut=off} % DO NOT CHANGE THIS
\usepackage{algorithm}
\usepackage{algorithmic}

\usepackage{newfloat}
\usepackage{listings}
\floatstyle{ruled}
\newfloat{listing}{tb}{lst}{}
\floatname{listing}{Listing}

\usepackage{amsthm,amsmath,amssymb}
\usepackage{mathrsfs}

\usepackage{epstopdf}

\usepackage{rotating}
\usepackage{multirow}

\usepackage{booktabs}

\usepackage{tablefootnote}

\usepackage{enumitem}

\title{Hierarchy-Aware T5 with Path-Adaptive Mask Mechanism for Hierarchical Text Classification}
\author {
    Wei Huang,\textsuperscript{\rm 1}
    Chen Liu,\textsuperscript{\rm 2}\thanks{Corresponding Auther.}
    Yihua Zhao,\textsuperscript{\rm 2}
    Xinyun Yang,\textsuperscript{\rm 2}
    Zhaoming Pan,\textsuperscript{\rm 2}
    Zhimin Zhang,\textsuperscript{\rm 2}\\
    Guiquan Liu\textsuperscript{\rm 1}
}
\affiliations {
    \textsuperscript{\rm 1} University of Science and Technology of China\\
    \textsuperscript{\rm 2} NetEase Media Technology (Beijing) Co., Ltd\\
    hw001@mail.ustc.edu.cn, \{liuchen5, hzzhaoyihua, yangxinyun, panzhaoming, zhangzhimin\}@corp.netease.com, gqliu@ustc.edu.cn
}

\usepackage{bibentry}
\begin{document}

\maketitle

\begin{abstract}
Hierarchical Text Classification (HTC), which aims to predict text labels organized in hierarchical space, is a significant task lacking in investigation in natural language processing. Existing methods usually encode the entire hierarchical structure and fail to construct a robust label-dependent model, making it hard to make accurate predictions on sparse lower-level labels and achieving low Macro-F1. In this paper, we propose a novel PAMM-HiA-T5 model for HTC: a hierarchy-aware T5 model with path-adaptive mask mechanism that not only builds the knowledge of upper-level labels into low-level ones but also introduces path dependency information in label prediction. Specifically, we generate a multi-level sequential label structure to exploit hierarchical dependency across different levels with Breadth-First Search (BFS) and T5 model. To further improve label dependency prediction within each path, we then propose an original path-adaptive mask mechanism (PAMM) to identify the label’s path information, eliminating sources of noises from other paths. Comprehensive experiments on three benchmark datasets show that our novel PAMM-HiA-T5 model greatly outperforms all state-of-the-art HTC approaches especially in Macro-F1. The ablation studies show that the improvements mainly come from our innovative approach instead of T5.
\end{abstract}

\section{Introduction}
Hierarchical text classification (HTC), where text labels are predicted within a hierarchical structure, is a challenging task that has not yet received due attention within the field of multi-label classification. HTC methods have been extensively applied in industry domains, e.g., news article classification \cite{AB2/GZC6PL_2008}, product classification in E-commerce \cite{Yu2018MultilevelDL}, bidding strategy in paid search marketing \cite{agrawal2013multi}.

\begin{figure}[t]
	\centering
	\includegraphics[width=0.95\columnwidth]{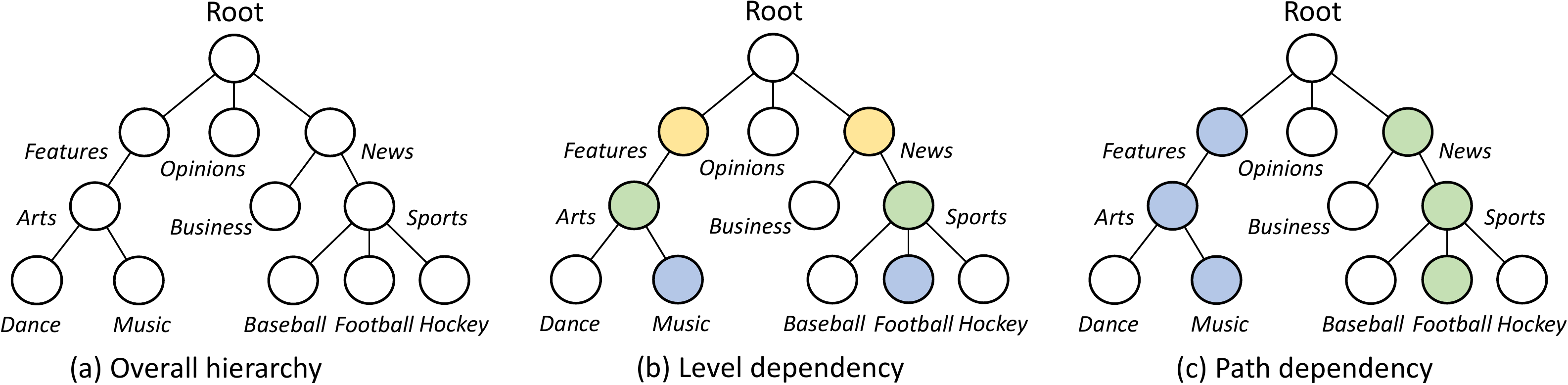} % Reduce the figure size so that it is slightly narrower than the column. Don't use precise values for figure width.This setup will avoid overfull boxes.
	\caption{(a): the static labeling process is uniform and simultaneous for all labels in the label hierarchy. (b): the dynamic labeling process where the lower-level labels depend on the upper-level labels. (c): the dynamic labeling process focuses on ancestor labels already generated on the current path. “Music” label is determined by “Arts” and “Features”, while “Football” label is ruled by “Sports”, “News”.}
	\label{label hierarchy}
\end{figure}

In HTC tasks, labels at lower-level are inevitably sparse due to the hierarchical structure. Many studies \citep{hayete2005gotrees,barbedo2006automatic,xiao2007hierarchical,johnson-zhang-2015-effective} completely or partially neglect such hierarchical structure and fail to accurately predict those lower-level labels, achieving low Macro-F1 score. Existing studies \cite{8933476,wu-etal-2019-learning-learn} have proved that introducing structure information can boost the predictive power on low-level labels and thus improve the overall task performance. A number of studies \citep{cesa2006hierarchical,shimura-etal-2018-hft,pmlr-v80-wehrmann18a,banerjee-etal-2019-hierarchical} propose to construct multi-level classifiers that are trained independently and predicted sequentially, where only local maximum is achieved and propagation of error negatively impacts model prediction. Some studies design an end-to-end model that introduces various strategies (such as Tree-LSTM/GCN \cite{zhou-etal-2020-hierarchy}, graph-CNN \cite{10.1145/3178876.3186005} and hierarchical fine-tuning based CNN \cite{shimura-etal-2018-hft}) to encode the overall hierarchy information (as depicted in Figure \ref{label hierarchy} (a)) and predict all labels simultaneously and independently with sigmoid function, where label dependency across different levels (as depicted in Figure \ref{label hierarchy} (b)) is not captured in a more principled way and unnecessary noises are introduced. Although one recent study \cite{mao-etal-2019-hierarchical} develops label-dependent models with reinforcement learning, it still fails to address label dependency within each path (as depicted in Figure \ref{label hierarchy} (c)) and fails to fully integrate labels and text information.

This paper seeks to close the gap by proposing the PAMM-HiA-T5. We are not only the first to capture lower-level label dependency on upper-level ones with generation model, but also the first to identify hierarchical dependency within the specific path. In each step of prediction phase, our model predicts next label based on the text sequence and labels previously generated on current path. As illustrated in Figure \ref{label hierarchy}(c), our model sequentially predicts “Features”, “News”, “Arts”, “Sports”, “Music”, “Football” labels. In the process where label “Music” is generated, our model pays more attention on “Features” and “Arts” labels on its own path instead of “News” and “Sports” labels on another path.

Our PAMM-HiA-T5 method follows a two-step design. 

\textbf{Hierarchy-aware T5 (HiA-T5)}, a variant of T5 that is fully aware of the level dependency. We firstly use Breadth-First Search (BFS) to flatten hierarchical labels into multi-level sequential label structure, transforming the hierarchy to sequence. T5 model is applied to map the text sequence to label sequence, where the text sequence and upper-level labels generated earlier are then integrated in order to determine the next label. As such, we are able to fully exploit label dependency across different levels. 

\textbf{Path-adaptive mask mechanism (PAMM)}, a mechanism to exploit the label correlation within each path and eliminate the noise of other paths. We propose the path-adaptive mask mechanism to separate the sources of noises from different paths. Regularization is introduced in the training phase to encourage the model to pay more attention to ancestor labels on current path while penalizing those on other paths, according to both the text sequence and labels generated.  

This study makes the following major contributions:
\begin{itemize}
	\item We propose a novel HiA-T5 model, a multi-level sequential label generative model to exploit label dependency across different levels. The mapping relationship between text sequence and label sequence is examined in each step of prediction.
	\item We propose an original PAMM to identify the label’s path information, separating the sources of noises from different paths to further improve prediction accuracy.
	\item Experiments on various datasets show that our PAMM-HiA-T5 model achieves significantly and consistently better performance than state-of-the-art models. Boosting Micro-F1 score by $2.29\%$ and achieving substantial $4.68\%$ improvement in Macro-F1 score, we establish new state-of-art results on RCV1-V2. The ablation studies show that the improvements mainly come from our innovative approach instead of T5.
\end{itemize}

\section{Related Work}
Hierarchical text classification (HTC) is a critical task with numerous applications \cite{qu2012evaluation,agrawal2013multi,zhang2019higitclass,peng2016deepmesh}. By methods of hierarchical information modeling, HTC approaches can be categorized into flat, local and global approaches \citep{silla2011survey}.

Flat approaches \citep{hayete2005gotrees,barbedo2006automatic,xiao2007hierarchical,johnson-zhang-2015-effective} completely or partially ignore the label hierarchy and each label is independently predicted. Some of them simply ignore the invaluable hierarchical information and achieve poor performance. Some others predict leaf nodes first and then mechanically add their ancestor labels, which is only applicable where different paths in the label hierarchy share the same length. 

Local approaches \citep{koller1997hierarchically,cesa2006hierarchical,shimura-etal-2018-hft,pmlr-v80-wehrmann18a,banerjee-etal-2019-hierarchical} construct multiple local classifiers so that the misclassification at a certain level is propagated downwards the hierarchy, easily leading to the exposure of bias \citep{silla2011survey}. Specifically, \citet{10.1145/1089815.1089821} proposes a top-down variant of SVM for HTC. \citet{10.1145/3178876.3186005} uses deep graph convolutional neural networks with hierarchical regularization. \citet{pmlr-v80-wehrmann18a} utilizes a multi-label neural network architecture with local and global optimization. To address the lower-level labels sparsity problem, \citet{shimura-etal-2018-hft} takes advantage of a CNN-based model with the fine-tuning method. \citet{banerjee-etal-2019-hierarchical} proposes to transfer the parameters of parent classifiers to initialize child classifiers for HTC task.

Global approaches \cite{10.1145/2487575.2487644,mao-etal-2019-hierarchical,wu-etal-2019-learning-learn,zhou-etal-2020-hierarchy,8933476}, where the entire structural information is encoded and all labels are simultaneously predicted, has become recent mainstream due to its better performance. \citet{10.1145/2487575.2487644} utilizes regularization to modify the SVM. Neural network architectures are also applied in global approaches. \citet{mao-etal-2019-hierarchical} handles HTC task with reinforcement-learning-based label assignment method. \citet{wu-etal-2019-learning-learn} uses meta-learning to model the label interaction for multi-label classification. \citet{zhou-etal-2020-hierarchy} utilizes the Bi-TreeLSTM and GCN to model hierarchical relationship and makes flat predictions for hierarchical labels. \citet{8933476} combines CNN, RNN, GCN, and CapsNet to model hierarchical labels. Although recent researchers have managed to introduce hierarchical information in different fashions, most of them still regard flat multi-label classification as the backbone of HTC where all labels are predicted simultaneously and independently. Their exploitation of hierarchical structure is far from insufficient.

\section{Problem Definition}
For HTC, we define the overall label hierarchy as a tree-like structure, denoted by $T=(L,E)$, where $L=\{l_1,l_2,\ldots,l_K\}$ refers to the set of all label nodes in the corpus and $K$ is the total number of them. $E$ refers to the set of edges indicating the nodes' parent-child relations. Formally, we denote text objects as $\mathcal{X}=\{X_1,X_2,\ldots,X_N\}$ and their labels as $\mathcal{L} =\{L_1, L_2,\ldots,L_N\}$. 

Each text object is represented by a text sequence $X_i=[x_1,x_2,\ldots,x_J]$, where $x_j$ is a word and $J$ is the number of words in the text object. Meanwhile, each text object $X_i$ is mapped to a original label set $L_i=\{l_1,l_2,\ldots,l_k,1\leq k< K\}$ that contains multiple labels. We then define a set of special symbols $S=\{\_,/,EOS\}$ to identify special hierarchical relationships in the hierarchy. 

All labels $L$ in the corpus constitute the overall label hierarchy $T$. The original label set $L_i=\{l_1,l_2,\ldots,l_k,0\leq k< K\}$ of any text object $X_i$ constitute an partial label hierarchy $T_i$ and $T_i\subset T$. We aim to train a model to predict corresponding label set $L_i$ for each text object $X_i$, where the label set $L_i$ are constrained by the hierarchy $T_i$.

\section{Background}
\begin{figure}[htbp]
	\centering
	\includegraphics[width=0.95\columnwidth]{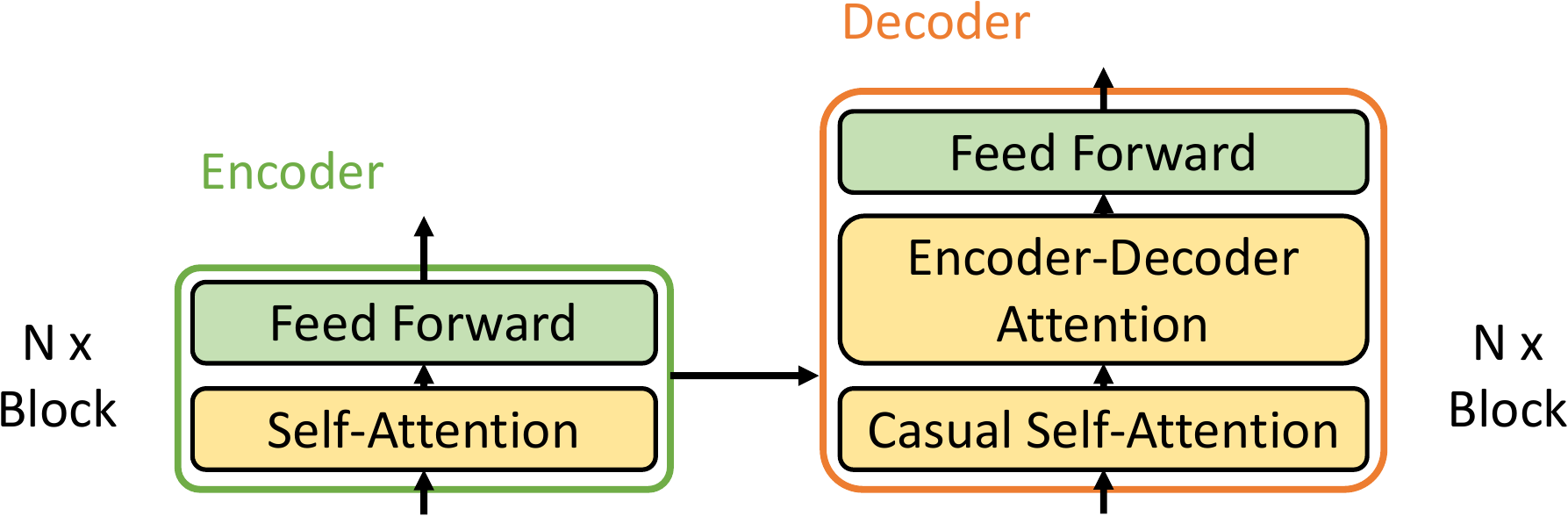} % Reduce the figure size so that it is slightly narrower than the column. Don't use precise values for figure width.This setup will avoid overfull boxes.
	\caption{Structure of T5. Following each Multi-head Attention sublayer and Feed-Forward sublayer, there are a series of dropout, residual connection and layer normalization. These parts are omitted in the figure and the following formulas for simplicity’s sake.}
	\label{T5 structure}
\end{figure}

The T5 model consists of an encoder-decoder architecture, which mainly includes the Multi-head Attention Mechanism, the Feed-Forward Network and so on \cite{JMLR:v21:20-074}, as depicted in the Figure \ref{T5 structure}.

Attention is calculated as:
\begin{equation}
Attention(Q,K,V)=Score(Q,K)V
\end{equation}
\begin{equation}
Score(Q,K)=softmax(\frac{QK^T}{\sqrt{d_k}})\label{eq2}
\end{equation}
where $Q,K,V\in\mathbb{R}^{n\times d_{model}}$ and the length of sequence is $n$. The attention score matrix $Score\in\mathbb{R}^{n\times n}$ is applied to the matrix $V$ to calculate the weighted sum and obtain the final attention result. The results upon independently executing attention mechanism on $H$ heads are concatenated to get Multi-head Attention.
\begin{equation}
MultiHead(Q,K,V)=Concat(head_1,\ldots,head_h)W^O
\end{equation}
\begin{equation}
head_i=Attention(QW_i^Q,KW_i^K,VW_i^V)
\end{equation}
where the projections matrices $W_i^Q,W_i^k,W_i^v\in\mathbb{R}^{d_{model}\times d_{head}}$ and $W^O\in\mathbb{R}^{d_{model}\times d_{model}}$ are learnable parameters. $d_{head}$ is the dimension of each head and $d_{model}=H\times d_{head}$.

The Feed-Forward Network consists of two linear transformations with a nonlinear activation function in between.
\begin{equation}
FFN(x) = max(0,xW_1+b_1)W_2+b_2
\end{equation}

T5 encoder is composed of a stack of ``encoder blocks'' and we define the number of blocks as $B$. Each block contains a self-attention sublayer and a feed-forward sublayer. The input sequence of encoder is mapped to the embedding $Q_{encoder},K_{encoder},V_{encoder}\in\mathbb{R}^{n\times d_{model}}$, which are then passed into the encoder. 
\begin{equation}
\begin{aligned}
&Block_{Encoder}(Q_{encoder},K_{encoder},V_{encoder})\\=&FFN(MultiHead(Q_{encoder},K_{encoder},V_{encoder}))
\end{aligned}
\end{equation}
\begin{equation}
\begin{aligned}
&Encoder(Q_{encoder},K_{encoder},V_{encoder})\\=&stack(Block_{Encoder}(Q_{encoder},K_{encoder},V_{encoder}))
\end{aligned}
\end{equation}

The structure of the decoder looks similar to that of the encoder, except that it has an additional encoder-decoder attention sublayer that attends to the output of the encoder stack, following each casual self-attention sublayer. In the end, we obtain the decoder output denoted as $O_{decoder}$.

\begin{equation}
\begin{aligned}
&Block_{Decoder}(Q_{decoder},K_{decoder},V_{decoder},O_{encoder})\\& =FFN(MultiHead(MultiHead(Q_{decoder},K_{decoder},\\& \quad ~ V_{decoder}),O_{encoder},O_{encoder}))
\end{aligned}
\end{equation}
\begin{equation}
\begin{aligned}
&Decoder(Q_{decoder},K_{decoder},V_{decoder},O_{encoder})\\&=stack(Block_{Decoder}(Q_{decoder},K_{decoder},V_{decoder},\\& \quad ~ O_{encoder}))
\end{aligned}
\end{equation}

\section{Hierarchy-Aware T5 with Path-Adaptive Mask Mechanism}
\begin{figure*}[t]
	\centering
	\includegraphics[width=0.9\textwidth]{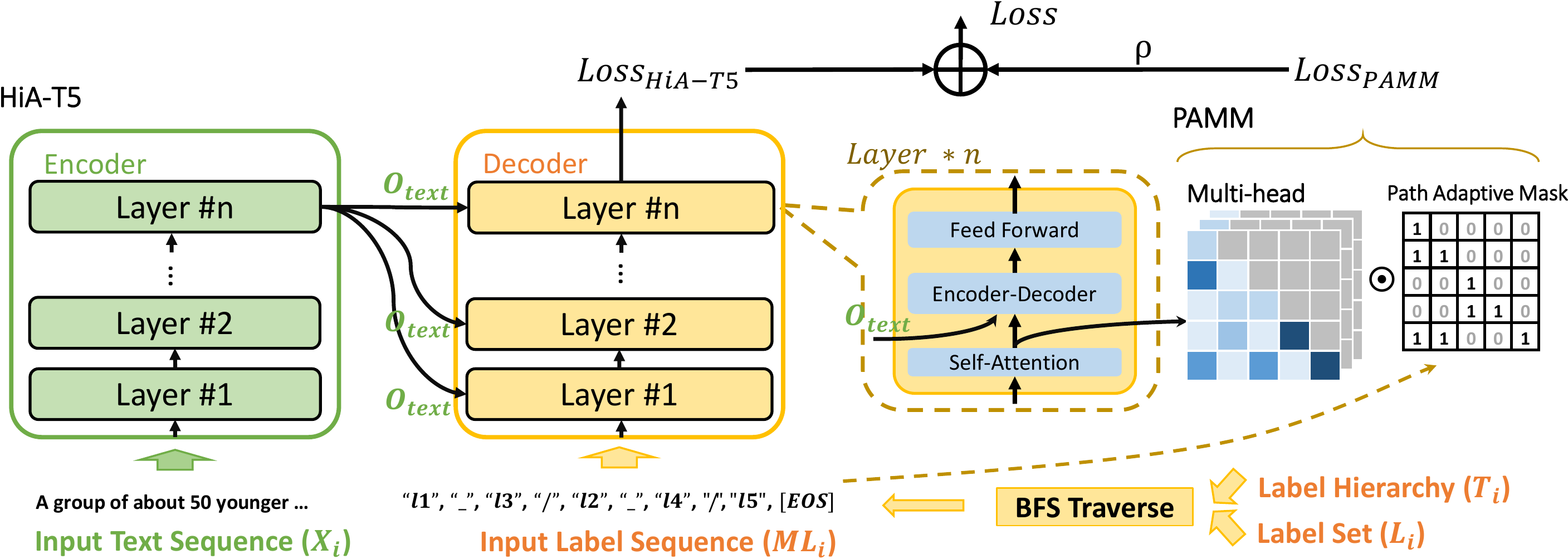} % Reduce the figure size so that it is slightly narrower than the column.
	\caption{The overall structure of PAMM-HiA-T5. PAMM-HiA-T5 consists of a HiA-T5 and a PAMM. The dataflows of one decoder layer are illustrated in the yellow dashed box.}
	\label{overall structure}
\end{figure*}

As depicted in Figure \ref{overall structure}, we propose a PAMM-HiA-T5 model for HTC: a \textbf{Hi}erarchy-\textbf{A}ware \textbf{T5} model with \textbf{P}ath-\textbf{A}daptive \textbf{M}ask \textbf{M}echanism. PAMM-HiA-T5 consisits of the HiA-T5 for level-dependent label generation and the PAMM for path-specific label generation.

\subsection{Hierarchy-Aware T5}

\paragraph{Level-dependent HiA-T5}

The major shortcoming of previous HTC methods is the inadequate application of hierarchy information. In contrast, HiA-T5 exploits label dependency across different levels of the hierarchy with Breadth-First Search (BFS) and multi-head attention mechanism.

\begin{figure}[htbp]
	\centering
	\includegraphics[width=0.95\columnwidth]{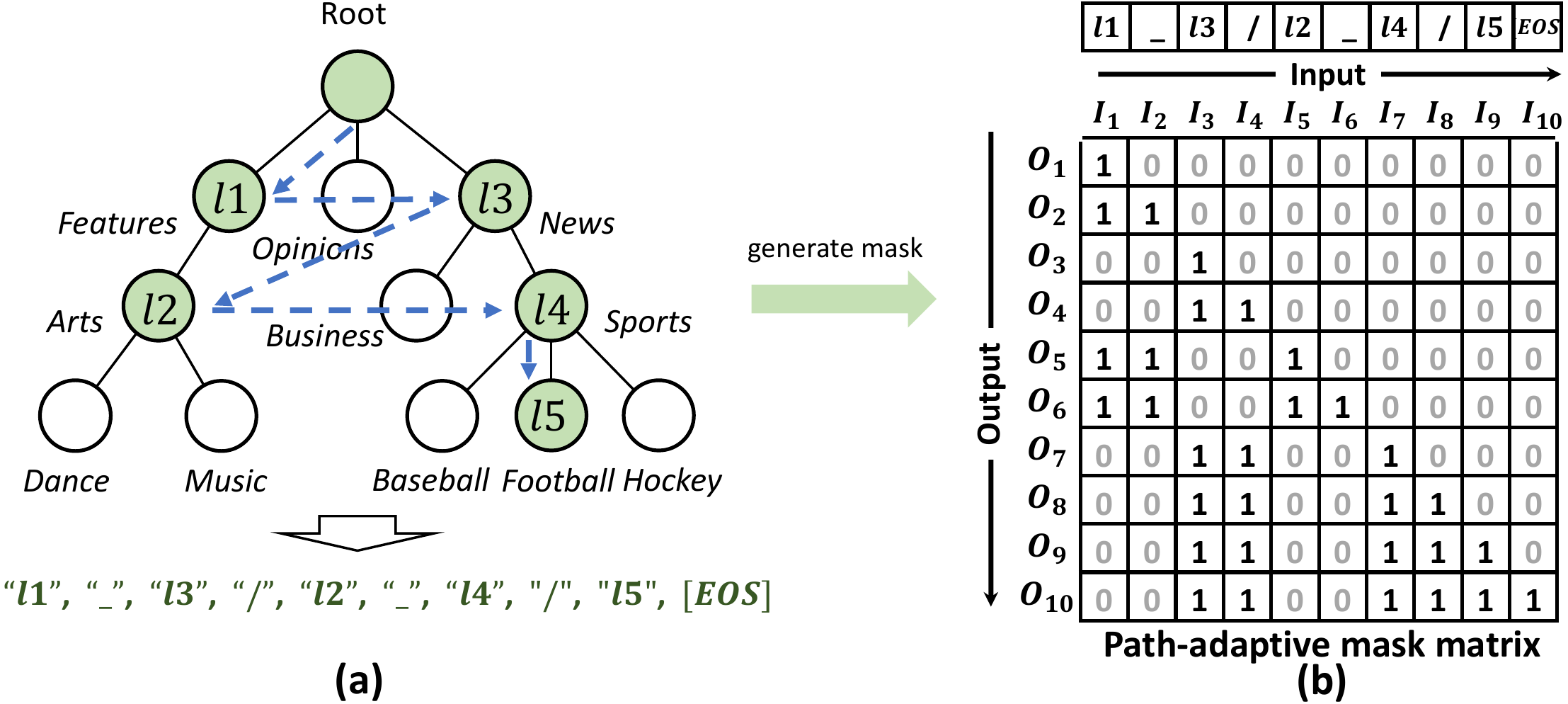} % Reduce the figure size so that it is slightly narrower than the column. Don't use precise values for figure width.This setup will avoid overfull boxes.
	\caption{(a): The label hierarchical structure is explored in Breadth-First Search (blue dash line). (b): path-adaptive mask matrix makes the $ith$ output element use current input element and all its ancestors.}
	\label{bfs2mask}
\end{figure}

HiA-T5 firstly explore the label hierarchy $T_i$ with Breadth-First Search \cite{cormen2001introduction} to flatten the label set $L_i=\{l_1,l_2,l_3,l_4,l_5\}$ into multi-level sequential label $ML_i=[l_1,\_,l_3,/,l_2,\_,l_4,/,l_5,EOS]$, transforming the hierarchy to multi-level label sequence, as illustrated in Figure \ref{bfs2mask} (a). In this process, ‘\_’ between labels denotes intra-level relationship, while ‘/’ signifies inter-level relationship.  

On one hand, the text sequence $X_i=[x_1,x_2,\ldots,x_J]$ is mapped to embedding sequence $Q_{text},K_{text},V_{text}\in\mathbb{R}^{n\times d_{model}}$, which are then passed into T5 encoder:
\begin{equation}
O_{text}=Encoder(Q_{text},K_{text},V_{text})
\end{equation}
The output encoder representation for semantic features of varied granularities is $O_{text}$.

On the other hand, the multi-level label sequence $ML_i=[l_1,\_,l_3,/,l_2,\_,l_4,/,l_5,EOS]$ is mapped to embeddings sequence $Q_{label},K_{label},V_{label}\in\mathbb{R}^{n\times d_{model}}$, which are passed into T5 decoder together:
\begin{equation}
O_{hierarchy}=Decoder(Q_{label},K_{label},V_{label},O_{text})
\end{equation}
Specifically, HiA-T5 fully explores the label dependency across different levels through the self-attention mechanism. With the help of the intra-level separator '\_' and the inter-level separator '/', the causal decoder self-attention mechanism fully excavates the intra-level parallel and mutually exclusive relationship, as well as the inter-level dependent and appurtenant relationship. The output representation of the decoder causal self-attention mechanism incorporating level dependency information is $A_{label}=MultiHead(Q_{label},K_{label},V_{label})$.

So far, we have obtained the text representation $O_{text}$ highlighting the semantic features of texts with different granularities and the label representation $A_{label}$ incorporating label dependency across different levels. The output representation of the encoder-decoder attention mechanism integrating these two is $A_{cross}=MultiHead(A_{label},O_{text},O_{text})$, which is a sufficient crossover information for following prediction.

\paragraph{Loss of HiA-T5}

We have obtained the final decoder block output of HiA-T5 $O_{hierarchy}$, which fully integrates the label hierarchy information and the text semantic information of different granularities. Then $O_{hierarchy}$ is passed into a fully connected layer with a softmax output, which is also the final result of HiA-T5 denoted as $Pred$. $Pred$ is the result of $n$ timesteps and $Pred\in\mathbb{R}^{n\times K}$.
\begin{equation}
\begin{aligned}
Pred=softmax(O_{hierarchy}W_3+b_3)
\end{aligned}
\end{equation}
where $W_3\in\mathbb{R}^{d_{model}\times K},b_3\in\mathbb{R}^{K}$. In addition, any multi-level label sequence $ML_i$ is transformed into $Truth\in\mathbb{R}^{n\times K}$, which is composed of one-hot vectors corresponding to all labels. Therefore, the cross-entropy loss of HIA-T5 expressed is as follows:
\begin{equation}
\begin{aligned}
%Loss_{HiA-T5}=-\sum_{i=1}^{n}(\sum_{k=1}^{K}(truth_{i,k}log(pred_{i,k})))
Loss_{HiA-T5}=crossentropy(Truth,Pred) 
\end{aligned}
\end{equation}

\subsection{Path-Adaptive Mask Mechanism}
PAMM is the regularization designed in the training phase to encourage the model to pay more attention to ancestor labels on current path while penalizing those on other paths. We first obtain the path-adaptive mask matrix containing hierarchy information. Then path-adaptive mask loss is obtained according to operations on the path-adaptive mask matrix and the causal attention score matrix.
\paragraph{Path-Adaptive Mask Matrix}
Now the text sequence $X_i=[x_1,x_2,\ldots,x_J]$ is taken as input, which is fed into HiA-T5 for training, and its corresponding multi-level label sequence $ML_i$ is taken as output.

\begin{figure}[htbp]
	\centering
	\includegraphics[width=0.95\columnwidth]{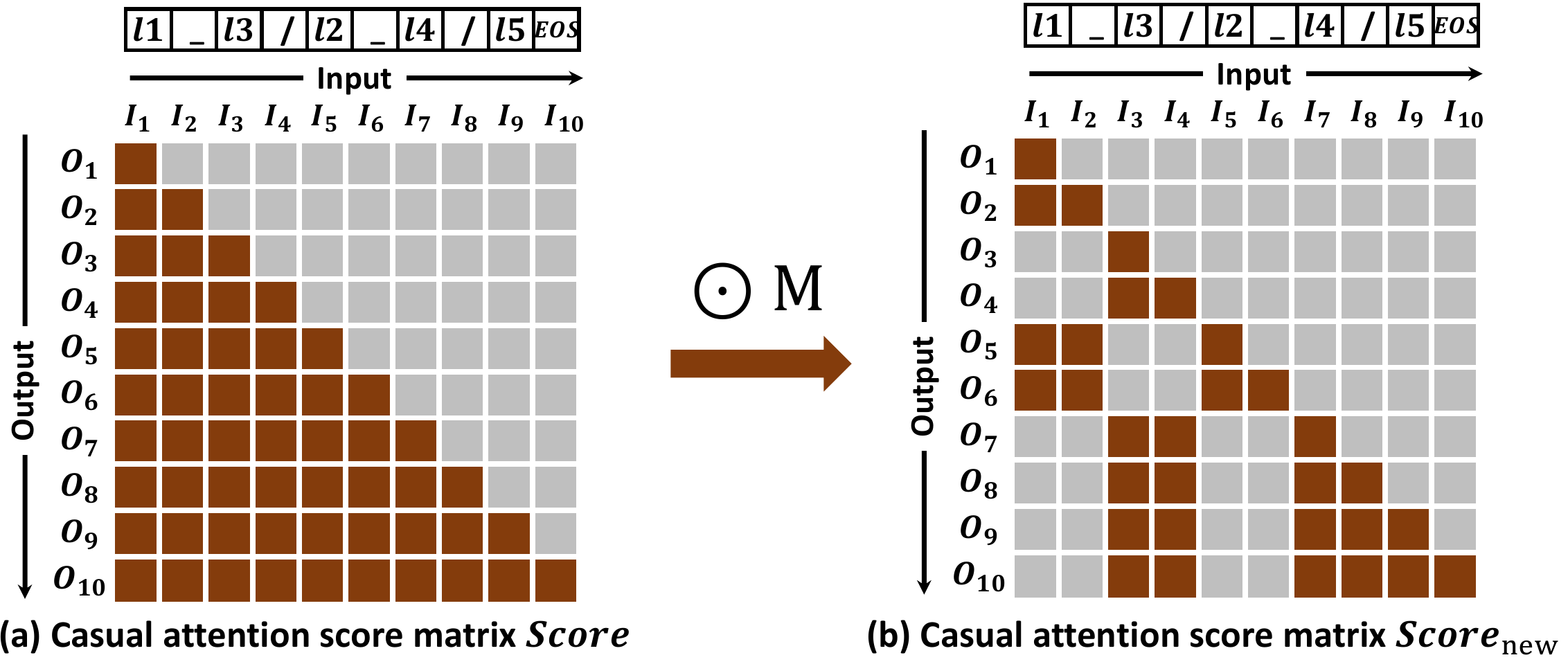} % Reduce the figure size so that it is slightly narrower than the column. Don't use precise values for figure width.This setup will avoid overfull boxes.
	\caption{(a): Causal attention score matrix $Score$. The input and output of the causal self-attention mechanism are denoted as $I$ and $O$ respectively. $Score=\left\{s_{i,j}\right\}\in\mathbb{R}^{n\times n}$. Each element $s_{i,j}$ at row $i$ and column $j$ represents the weight at which the self-attention mechanism attends to input element $j$ at output timestep $i$. The gray cell indicates the corresponding attention score $s_{i,j}=0$. (b): Element-wise product result of $Score$ and $M$.}
	\label{attention score}
\end{figure}

According to the T5 structure of Figure \ref{T5 structure}, the sequence $ML_i$ is first passed into causal attention sub-layer of decoder. Within this sub-layer, and according to formula (\ref{eq2}), we get causal attention score matrix $Score$ corresponding to the sequence $ML_i$, as depicted in Figure \ref{attention score} (a).

Then we define the path-adaptive dynamic mask matrix $M$, which can mask different parts of the label sequence at different decoding timestep $i$. The matrix $M$ is obtained from the hierarchical structure of the label sequence corresponding to each text object. Specifically, the shape of the matrix $M$ is same as the causal attention score matrix $Score$. Mask matrix $M$ is also a lower triangular matrix and $M=\{m_{i,j}\}\in\mathbb{R}^{n\times n}$, which is only composed of 0 or 1 as shown below.

$
M=\begin{bmatrix}
m_{1,1} &  &  &  & 0\\
m_{2,1} & m_{2,2} & \\
\vdots & \vdots & \ddots & \ddots\\
m_{n,1} & m_{n,2} & \cdots & m_{n,n-1} & m_{n,n}\\
\end{bmatrix}
$

$I_i$ represents the input of the attention mechanism at the timestep $i$, and $I_i \in L\cup S$. If $I_i\in L$, we define $ancestor(I_i)$ as label $I_i$'s ancestor labels and the special symbol immediately following it.

Then we define the following formula to fill the matrix $M$ based on the parent-child relationship contained in each path of label hierarchy. The $i$-th input timestep of the causal attention mechanism corresponds to the $i$-th row of matrix $M$, which contains $i$ elements: $m_{i,1},m_{i,2},\ldots,m_{i,i}$. In the $i$-th row of matrix $M$, inputs $I_1,I_2,\ldots,I_i$ of different timesteps corresponds to elements $m_{i,1},m_{i,2},\ldots,m_{i,i}$ respectively. 

Note that, in the $i$-th ($1\leq i\leq n$) row of the matrix $M$, we first determine the value of the diagonal element $m_{i,i}$, then determine the other elements $m_{i,j}(1\leq j<i)$ before $m_{i,i}$, and the formula is as follows:

\begin{equation}
\resizebox{\linewidth}{!}{$
	m_{i,j}=\left\{
	\begin{array}{rl}\displaystyle
	1  & {\{I_i\in L, I_j\in ancestor(I_i), 1\leq j<i\}}\\\displaystyle
	&  \cup{\{I_i\in S, j=i-1\}}\\\displaystyle
	&  \cup{\{I_i\in S, I_j\in ancestor(I_{i-1}), 1\leq j<i\}}\\\displaystyle
	0  & {else}
	\end{array} \right.$}
\end{equation}

\paragraph{Path-Adaptive Mask Loss}

With text sequence and labels previously generated in hand, we now introduce regularization and apply the path-adaptive dynamic mask matrix $M$, such that HiA-T5 decoder learns the weight of the attention matrix and pays more attention on the label’s current path.

Having obtained the multi-level label sequence $ML_i$ of a certain training sample, we use it as the input of causal self-attention of HiA-T5's decoder. According to the definition above, we get its path-adaptive mask matrix $M$, as depicted in Figure \ref{bfs2mask} (b). Furthermore, we get a new attention score matrix $Score_{new}$ as depicted in Figure \ref{attention score} (b) by multiplying attention score matrix $Score$ and the mask matrix $M$ element-wise:

\begin{equation}
Score_{new} = Score\odot M=softmax(\frac{QK^T}{\sqrt{d_k}})\odot M
\end{equation}

We define $C$ as the index set of $ancestor(I_i)$. At any decoding timestep $i$, our goal is to make the sum of the attention scores $\sum_{j\in C}s_{i,j}$ of current path's labels as close to $1$ as possible. Corresponding to attention scores matrix $Score_{new}$ of decoder's causal attention, that is, to make the sum of elements of each row in the matrix close to $1$ as much as possible. According to the definition section, suppose $Score$ is the causal attention score matrix corresponding to the $h$-th head of $b$-th decoder ``blocks'', where $1\leq h\leq H,1\leq b\leq B$. The path-adaptive mask loss is defined as:

\begin{equation}
\begin{aligned}
Loss_{PAMM}=\sum_{b=1}^{B}(\frac{\sum_{h=1}^{H}(\sum_{i=1}^{n}(1-\sum_{j\in C}s_{i,j}))}{H})
\end{aligned}
\end{equation}

Therefore, the loss generated by the path adaptive mask mechanism is added to the loss of HiA-T5 as total loss for training. The total loss function $Loss$ is obtained as below, where $\rho$ is the coefficient of path-adaptive mask loss item.
\begin{equation}
	\begin{aligned}
	%Loss=&Loss_{HiA-T5}+\rho Loss_{PAMM}\\=&-\sum_{i=1}^{n}(\sum_{k=1}^{K}(truth_{i,k}log(pred_{i,k})))\\&+\rho\sum_{b=1}^{B}(\frac{\sum_{h=1}^{H}(\sum_{i=1}^{n}(1-\sum_{j\in C}a_{i,j}))}{H})
	Loss=&Loss_{HiA-T5}+\rho Loss_{PAMM}
	\end{aligned}
\end{equation}

\section{Experiments}

\subsection{Experiment Setup}

\paragraph{Datasets}

We conduct extensive experiments on three public datasets, including RCV1-V2 \cite{lewis2004rcv1}, NYTimes(NYT) \cite{AB2/GZC6PL_2008} and Web-of-Science(WOS) \cite{kowsari2017hdltex}. RCV1-V2 and NYT are both news categorization dataset while WOS is about scientific literature categorization. Labels of these datasets are organized into a tree-like structure. Relevant information of datasets is summarized in Table \ref{tab1} and Table \ref{tab_level}.

We split RCV1-V2 in the benchmark dataset split manner and take a small portion of the training set as validation set. For NYT and WOS, we randomly split data into training, validation and test sets.

\begin{table}[h]
	\centering
	\resizebox{\columnwidth}{!}{
	\begin{tabular}{cccccccc}
		\hline
		Dataset & $\left| L \right|$ & Depth & Avg($\left| L_i \right|$) & Max($\left| L_i \right|$) & Train & Val & Test\\
		\hline
		RCV1 & 103 & 4 & 3.24 & 17 & 20833 & 2316 & 781265\\
		NYT & 166 & 8 & 7.6 & 38 & 23345 & 5834 & 7292\\
		WOS & 141 & 2 & 2.0 & 2 & 30070 & 7518 & 9397\\ 
		\hline
	\end{tabular}}
	\caption{Statistical analysis of datasets:  $\left| L_i \right|$ is the number of all labels in the hierarchy. Depth denotes the maximum level of the label hierarchy. Avg($\left| L_i \right|$) and Max($\left| L_i \right|$) denote average and maximum number of labels in each sample.}
	\label{tab1}
\end{table}

\begin{table}[h]
	\centering
	\resizebox{\columnwidth}{!}{
	\begin{tabular}{ccccccccc}
		\hline
		Dataset & level1 & level2 & level3 & level4 & level5 & level6 & level7 & level8\\
		\hline
		RCV1 & 236334 & 20523 & 11850 & 23211 & - & - & - & - \\
		NYT & 15161 & 2923 & 1160 & 842 & 1066 & 925 & 992 & 1460\\
		WOS & 6712 & 351 & - & - & - & - & - & -\\
		\hline
	\end{tabular}}
	\caption{Statistics of the average number of each label’s occurrence at each level: $level_i$ denotes the level in the label hierarchy. In general, lower-level labels are more sparse.}
	\label{tab_level}
\end{table}

\paragraph{Evaluation Metrics}

We use standard evaluation metrics, including \textbf{Micro-F1} and \textbf{Macro-F1} \cite{10.1145/2487575.2487644,10.1145/3178876.3186005,10.1145/3357384.3357885}, to measure the performance of all HTC methods. Micro-F1 equally weights all samples, while Macro-F1 gives equal weight to each label. As such, Micro-F1 gives more weight to frequent labels, while Macro-F1 equally weights all labels and is more sensitive to lower-level labels which are hard to predict. As shown in Table \ref{tab_level}, the labels of most samples are screwed towards upper levels, in which case Macro-F1 is more sensitive to scattered lower-level labels. 

\paragraph{Experimental Settings}

The backbone pre-trained model we adopt is T5-base \cite{JMLR:v21:20-074}. T5-base contains a total of about 220M parameters, including 12 layers of transformer, and each layer has 768 hidden dimensions, 3072 intermediate sizes and 12 attention heads. T5 is fine-tuned on relevant datasets before we perform following experiments. For the reproducibility of results, we set random seeds before experiments. Tokenizer from T5 is utilized to preprocess the text. For T5, the maximum length of token inputs of encoder is set as $300$, and the maximum length of token outputs of decoder is set as $60$. When the T5 model is fine-tuned, Adam optimizer is employed in a mini-batch size of $10$ with learning rate $3\times10^{-4}$, and model converges after 3 epochs. The search range of coefficient $\rho$ is \{0.1,1,10,100,200\}, and we set it to $100$ according results of validation set. The model with the best performance on the validation set is selected for evaluating the test set. In the inference phase, greedy search is adopted. The experiments are conducted on GeForce RTX 2080 Ti GPU and CentOS with Pytorch Lightning framework.

\subsection{Performance Comparison} 

Experimental results on RCV1-V2 benchmark dataset are shown in Table \ref{tab2}, and our proposed PAMM-HIA-T5 outperforms all state-of-the-art results of flat, local and global approaches by a large margin, both in Micro-F1 and Macro-F1. This demonstrates the strong power of PAMM-HiA-T5 in solving HTC problems. PAMM-HIA-T5 achieve the performance of $86.25\%$ Micro-F1 score and $68.03\%$ Macro-F1 score, which outperforms state-of-the-art model by $4.68\%$ of Macro-F1 and $2.29\%$ of Micro-F1 on RCV1-V2. The greater improvement on Macro-F1 shows that PAMM-HiA-T5 has greater capability in predicting sparse lower-level labels, which results from the fact that our model utilizes the knowledge of upper-level labels in predicting lower-level ones by modeling hierarchical dependency. It can be shown from Table \ref{tab_level} that the samples’ labels are unevenly distributed across levels. The labels of most samples are screwed towards upper levels. As level grows, sample labels become more sparse. This explains the reason why our model achieves greater boost in Macro-F1 than Micro-F1.

Experiment results on WOS and NYT datasets are shown in Table \ref{tab3}, and we obtain consistent conclusion as from RCV1-V2: the level-dependency modeling and the path-adaptive mask mechanism bring significant performance improvement. On NYT dataset, we obtain the best performance of $77.89\%$ Micro-F1 and $65.13\%$ Macro-F1, improving Micro-F1 by $2.92\%$ and Macro-F1 by $4.30\%$ compared with the result reported in latest state-of-the-art HTC approach \cite{zhou-etal-2020-hierarchy}. On WOS dataset, we still outperform the result reported in latest state-of-the-art HTC approach \cite{zhou-etal-2020-hierarchy}, although the improvement is minor due to its simple hierarchical structure and fewer lower-level labels. 	

\begin{table}[h]
	\centering
	\resizebox{\columnwidth}{!}{
	\begin{tabular}{clcc}
		\cmidrule[1pt]{2-4} &\textbf{Model} & Micro-F1 & Macro-F1\\
		\cmidrule{2-4} \multirow{5}{*}{\rotatebox{90}{Flat}}
		&Leaf-SVM* & $69.05$ & $32.95$\\
		&SVM & $81.60$ & $60.70$\\
		&HAN** & $75.30$ & $40.60$\\
		&TextCNN** & $76.60$ & $43.00$\\
		&bow-CNN** & $82.70$ & $44.70$\\
		\cmidrule{2-4} \multirow{5}{*}{\rotatebox{90}{Local}}
		&TD-SVM** \cite{10.1145/1089815.1089821} & $80.10$ & $50.70$\\
		%&HSVM $^{*}$ & $69.3$ & $33.3$\\
		%&HR-SVM* & $72.8$ & $38.6$\\
		&HR-DGCNN-3 \cite{10.1145/3178876.3186005} & $76.18$ & $43.34$\\
		&HMCN \cite{pmlr-v80-wehrmann18a} & $80.80$ & $54.60$\\
		&HFT(M) \cite{shimura-etal-2018-hft} & $80.29$ & $51.40$\\
		&Htrans \cite{banerjee-etal-2019-hierarchical} & $80.51$ & $58.49$\\
		\cmidrule{2-4} \multirow{5}{*}{\rotatebox{90}{Global}}
		&HR-SVM*\tablefootnote{The results of HR-SVM reported in \cite{10.1145/2487575.2487644} are not comparable because they use a different hierarchy with 137 labels.} \cite{10.1145/2487575.2487644} & $72.75$ & $38.58$\\
		&SGM\tablefootnote{The result is reproduced with benchmark split upon the released project of SGM by \cite{zhou-etal-2020-hierarchy}.} \cite{yang-etal-2018-sgm} & $77.30$ & $47.49$\\
		&HE-AGCRCNN \cite{8933476} & $77.80$ & $51.30$\\
		&HiLAP-RL \cite{mao-etal-2019-hierarchical} & $83.30$ & $60.10$\\
		&HiAGM \cite{zhou-etal-2020-hierarchy} & $83.96$ & $63.35$\\
		&\textbf{PAMM-HiA-T5} & $\mathbf{86.25}$ & $\mathbf{68.03}$\\
		\cmidrule[1pt]{2-4}
	\end{tabular}}
	\caption{Performance comparison on RCV1-V2. $*$ denotes the results reported in \cite{10.1145/3178876.3186005} and $**$ denotes the results reported in \cite{mao-etal-2019-hierarchical} on the similar dataset split. In this table, SVM \cite{noble2006support} implements standard multi-label classification in a way of one-vs-the-rest(OVR). Leaf-SVM predicts leaf nodes first and then mechanically add their ancestor labels. TextCNN \cite{kim-2014-convolutional}, HAN \cite{yang-etal-2016-hierarchical} and bow-CNN \cite{johnson-zhang-2015-effective} are modified as flat baselines for HTC by \cite{mao-etal-2019-hierarchical}. This is the case for Table \ref{tab3} too.}
	\label{tab2}
\end{table}

\begin{table}[!h]
	\centering
	\resizebox{\columnwidth}{!}{
	\begin{tabular}{lcccc}
		\toprule \multirow{2}{*} { \textbf{Model} } & \multicolumn{2}{c} { NYT } &  \multicolumn{2}{c} { WOS } \\
		\cmidrule(r){2-3}\cmidrule(r){4-5}
		& Micro-F1 & Macro-F1 &  Micro-F1 & Macro-F1 \\
		\midrule 
		SVM* & $72.40$ & $37.10$ &  $-$ & $-$ \\
		TextCNN* & $69.50$ & $39.50$ &  $-$ & $-$ \\
		HAN* & $62.80$ & $22.80$ &  $-$ & $-$ \\
		bow-CNN* & $72.90$ & $33.40$  &  $-$ & $-$ \\
		TD-SVM* \cite{10.1145/1089815.1089821} & $73.70$ & $43.70$ &  $-$ & $-$ \\
		HMCN* \cite{pmlr-v80-wehrmann18a} & $72.20$ & $47.40$ &  $-$ & $-$ \\
		HiLAP* \cite{mao-etal-2019-hierarchical} & $74.60$ & $51.60$ &  $-$ & $-$ \\
		HiAGM \cite{zhou-etal-2020-hierarchy} & $74.97$ & $60.83$ & $85.82$ & $80.28$ \\
		\midrule  
		\textbf{PAMM-HiA-T5} & $\mathbf{77.89}$ & $\mathbf{65.13}$ &  $\mathbf{90.36}$ & $\mathbf{81.64}$ \\
		\bottomrule
	\end{tabular}}
	\caption{Performance comparison on the NYT and WOS datasets. We focus on the most competitive methods which achieved the latest state-of-art results, that is \cite{mao-etal-2019-hierarchical} and \cite{zhou-etal-2020-hierarchy}. For fair comparison, NYT and WOS datasets are selected and splited in exactly the same way as \cite{zhou-etal-2020-hierarchy} does. And $*$ denotes the results reported in \cite{mao-etal-2019-hierarchical} on the similar dataset split.}
	\label{tab3}
\end{table}

\subsection{Performance Analysis}

\begin{table}[t]
	\centering
	%\resizebox{.95\columnwidth}{!}{
	\begin{tabular}{lcc}
		\hline
		Method & Micro-F1 & Macro-F1\\
		\hline
		T5 & $84.93$ & $62.31$ \\
		HiA-T5 & $85.99$ & $65.16$ \\
		PAMM-HiA-T5 & $86.25$ & $68.03$ \\
		\hline
	\end{tabular}
	\caption{Ablation study of PAMM-HiA-T5. Note that original T5 neither model the hierarchical structure information nor capture the hierarchial dependencies. It takes HTC as a generic multi-label classification task to generate unordered label sets corresponding to the text.}
	\label{tab4}
\end{table}

\paragraph{Ablation Study on Level Dependency Modeling}

We compare the performance of HiA-T5 with the original T5 model, shown in the first two rows of Table \ref{tab4}. It is evident that HiA-T5 greatly outperforms the T5 both in Micro-F1 and Macro-F1, and the improvement in Macro-F1 is greater than that in Micro-F1. This result illustrates the effectiveness of capturing level dependency by introducing upper-level label knowledge to assist lower-level label prediction. 

Our ablation studies show that the improvements are mainly due to the strategy and the mechanism we proposed rather than the T5 itself. Experiments on RCV1-V2 show that PAMM-HiA-T5 outperforms the sota model by $4.68\%$ in Macro-F1 and $2.29\%$ in Micro-F1. But T5 only contributes $0.97\%$ of the growth in Micro-F1, and it doesn't even exceed the sota model in Macro-F1.

\begin{figure}[!h]
	\centering
	\includegraphics[width=0.9\columnwidth]{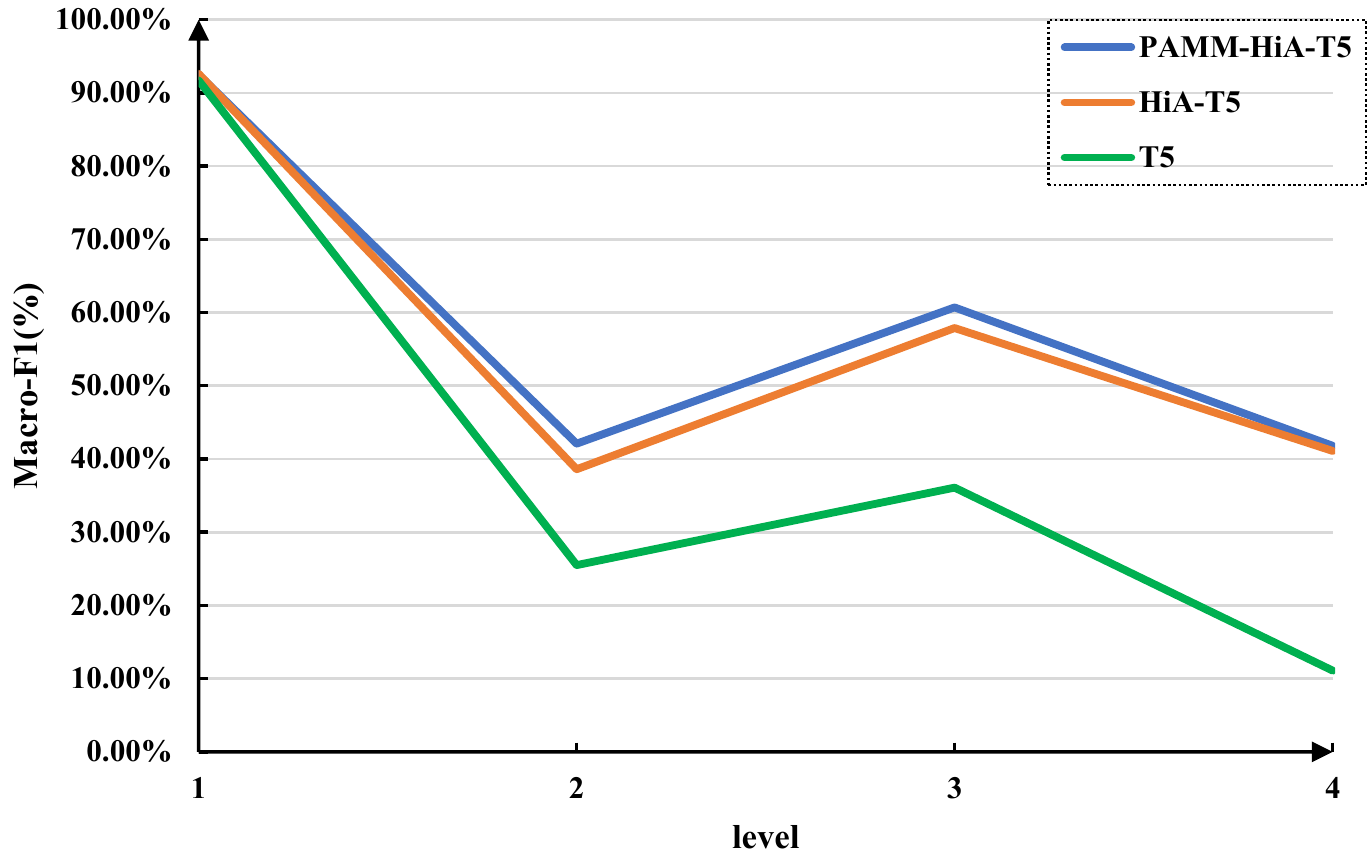} % Reduce the figure size so that it is slightly narrower than the column. Don't use precise values for figure width.This setup will avoid overfull boxes.
	\caption{Level-based Macro-F1 score on RCV1-V2.}
	\label{level statistics}
\end{figure}

In addition, Figure \ref{level statistics} demonstrates that the gap between HiA-T5 and T5 gets bigger as the level deepens. This illustrates that as the level grows, label prediction becomes more and more difficult, and the introduction of upper-level label knowledge by HiA-T5 becomes more and more valuable.

\paragraph{Ablation Study on Path-adaptive Mask Mechanism}

We compare the performance of PAMM-HiA-T5 with the HiA-T5, and the ablation results are shown in the last two rows of Table \ref{tab4}. PAMM-HiA-T5 improves Macro-F1 by $2.87\%$ and Micro-F1 by $0.26\%$ compared with HiA-T5. This indicates that PAMM significantly improves the performance of HiA-T5 in more challenging multi-path scenarios by capturing precise path dependencies. With the assistance of PAMM, PAMM-HiA-T5 further improves the overall performance of HTC task especially in Macro-F1 by exploiting label dependency within each path and eliminating the noise from other paths. 

\begin{figure}[htbp]
	\centering
	\includegraphics[width=0.7\columnwidth]{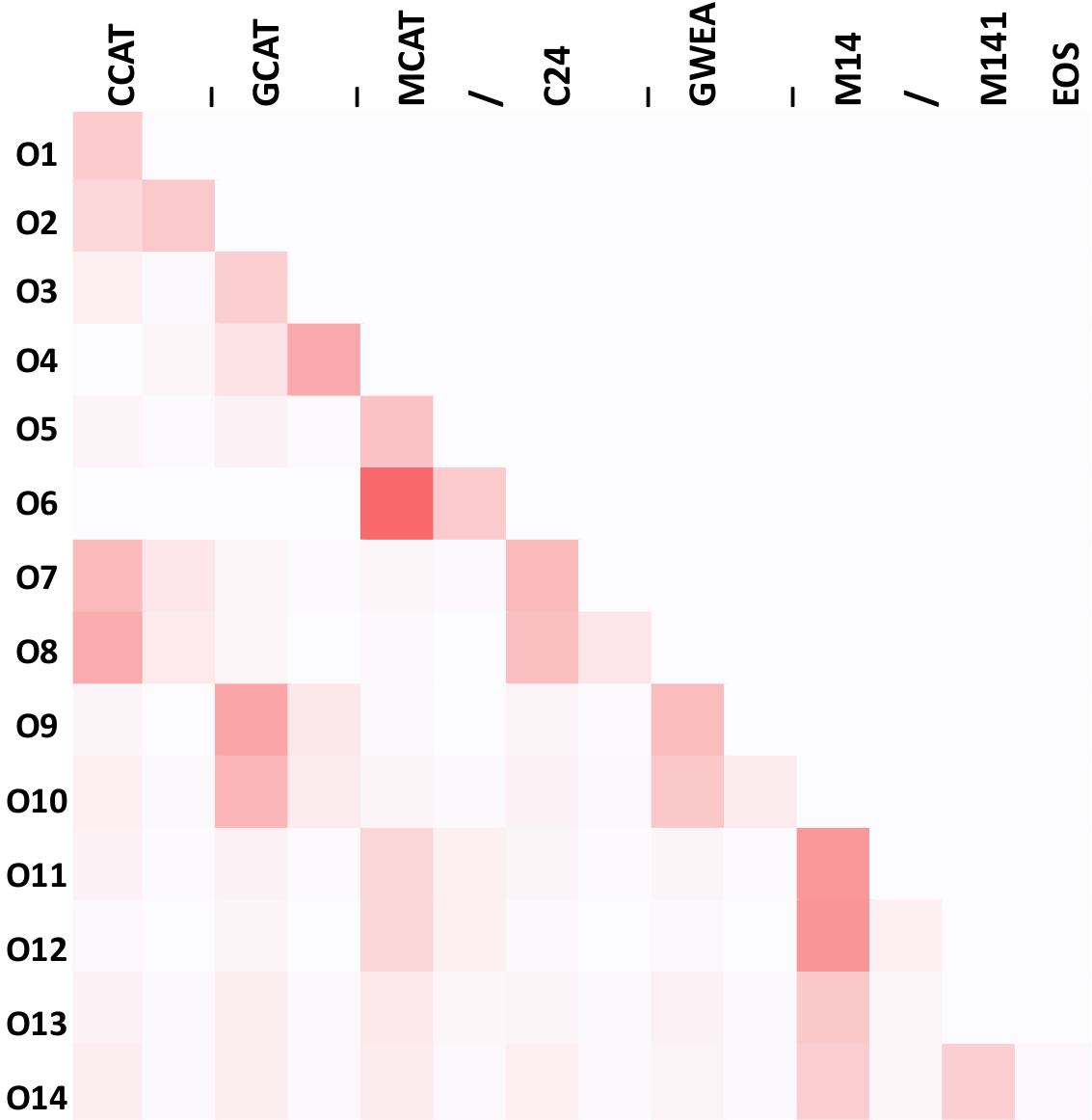} % Reduce the figure size so that it is slightly narrower than the column. Don't use precise values for figure width.This setup will avoid overfull boxes.
	\caption{Causal self-attention score's heat map corresponding to a random sample. We use symbols instead of original labels for ease of display. Note that the score of each label is the average of its tokens' attention score for a cleaner view.}
	\label{heat map}
\end{figure}

As shown in Figure \ref{heat map}, the heat map of the causal self-attention score in PAMM-HiA-T5's encoder proves the effectiveness of PAMM, where the attention score is mainly distributed on the path of the label current being decoded.

\paragraph{Analysis of Label Consistency}

Label inconsistency, where one label is predicted positive but its ancestors are not, is a serious problem in many HTC approaches, due to the fact that they focus on flat multi-label classification and make independent predictions for all labels. It is worth mentioning that PAMM-HIA-T5 has outstanding classification performance while maintaining an ultra-low label inconsistency rate of $0.31\%$, as shown in Table \ref{inconsistency rate}. This is because PAMM-HiA-T5 fully leverages the constraints of upper-level labels generated earlier to predict the most accurate lower-level labels.

\begin{table}[htbp]
	\centering
	\resizebox{.95\columnwidth}{!}{
	\begin{tabular}{cccc}
		\hline
		SVM & TextCNN & HMCN & \textbf{PAMM-HiA-T5}\\
		\hline
		$4.83\%$ & $3.74\%$ & $3.84\%$ & $\boldsymbol{0.31\%}$\\ 
		\hline
	\end{tabular}}
	\caption{Comparison of label inconsistency. The label inconsistency is calculated as the ratio of predictions with inconsistent labels. The results of SVM, TextCNN, and HMCN are reported in \citep{mao-etal-2019-hierarchical}.}
	\label{inconsistency rate}
\end{table}

\section{Conclusion}

This paper aims to improve HTC task performance by exhaustively exploring level and path dependency within the hierarchical structure. We have devised an innovative PAMM-HiA-T5 methodology in order to capture lower-level label dependency on upper-level ones with generation model and to identify hierarchical dependency within the specific path. In the first place, we generate a multi-level sequential label structure to exploit level-dependency with Breadth-First Search (BFS) and T5 model. To further capture label dependency within each path, we then propose an original path-adaptive mask mechanism (PAMM) to identify the label’s path information, eliminating sources of noises from other paths. Comprehensive experiments on three benchmark datasets show that our novel PAMM-HiA-T5 model establishes new state-of-the-art results, achieving significant Macro-F1 improvement by $4.68\%$ and Micro-F1 improvement by $2.29\%$. The ablation studies show that the improvements mainly come from our original approach instead of T5.

%添加reference
\bibliography{aaai22}

\end{document}